\documentclass[journal]{IEEEtran}
\usepackage[]{graphicx}
\usepackage{amsmath}
\usepackage{amsfonts}
\usepackage{subcaption}
\usepackage{float}

\ifCLASSINFOpdf
\else
\fi

\begin{document}
%
\title{Dictionary based Approach to Edge Detection}
%
%
%

\author{Nitish~Chandra*,~\IEEEmembership{}
        Kedar~Khare~\IEEEmembership{} \\
        Department of Physics, Indian Institute of Technology, Delhi.
\thanks{*Nitish Chandra, e-mail: nitishchandra209@gmail.com.}
\thanks{}
\thanks{}}

\maketitle

\begin{abstract}
Edge detection is a very essential part of image processing, as quality and accuracy of detection determines the success of further processing. We have developed a new self learning technique for edge detection using dictionary comprised of eigenfilters constructed using features of the input image. The dictionary based method eliminates the need of pre or post processing of the image and accounts for noise, blurriness, class of image and variation of illumination during the detection process itself. Since, this method depends on the characteristics of the image, the new technique can detect edges more accurately and capture greater detail than existing algorithms such as Sobel, Prewitt Laplacian of Gaussian, Canny method etc which use generic filters and operators. We have demonstrated its application on various classes of images such as text, face, barcodes, traffic and cell images. An application of this technique to cell counting in a microscopic image is also presented.
\end{abstract}


%
\IEEEpeerreviewmaketitle

\section{Introduction}
%
%
%
%
\IEEEPARstart{I}{mage} processing techniques to enhance the quality of images or to extract information from an image have become
ubiquitous, in imaging systems such as cameras, microscopes, etc. Edge detection is an essential component of image processing as edges play an important role in visual perception. Edges in images can be defined as the points where sharp changes of colour and/or intensity occur. Current edge detection techniques utilize differential operators or combination of filtering operations or results from various techniques are combined together. There are many standard types of edge detection techniques, like Sobel, Prewitt and Laplacian of Gaussian that detect any discontinuities in color or intensity from one pixel to another in an image. 

In general, there are two categories of edge detection techniques. First, gradient methods in which maximum and minimum in first derivative is considered and other is Laplacian method which searches for zero crossing of second derivative. In practice, edges get blurred and noisy and the level of contamination depends on illumination, focusing mechanism, and noise level in electrical components of imaging system. Second derivative methods are more sensitive to noise than first derivative. Thus, there are three fundamental steps that are performed during edge detection, image smoothing for noise removal, detection of edge points, and edge localization. In Sobel and Prewitt methods these steps are performed separately and the results depend on the accuracy of all the steps. However, in Laplacian of Gaussian smoothing step is incorporated and Canny method consists of all the steps.

All the above methods use generic operators and do not consider the image characteristics. In this paper, we demonstrate an edge detection method based on image characteristics using a dictionary based algorithm. This method is different from traditional methods as it incorporates features of a given image, rather than using a set of pre-defined functions. 

In section II, we review popular existing algorithms for edge detection to which the results are compared in later sections. In section III, we present our algorithm of dictionary based edge detection. In section IV, we have compared the results with the popular techniques for different classes of images. We have also presented the application of the new method in barcode reading, counting of red blood cells and the results are compared with existing algorithms. We conclude and discuss further applications in section V.

\section{Review of Edge Detection Techniques}
Some of the most common techniques for edge detection employ  gradient operators i.e filtering the image with one or more masks. These methods do not have any provisions for image characteristics and noise content. Methods such as Sobel and Prewitt have gradient operator in orthogonal directions. For continuous function these are expressed as, 
\begin{equation}
\nabla f(x,y) = [G_x \hspace{5pt} G_y]^T = \big [\frac{\partial f}{\partial x} \hspace{5pt} \frac{\partial f}{\partial y}\big ]
\label{eq:grad}
\end{equation}

In practice for digital image processing calculation of partial derivatives, convolution is performed using a small kernel or mask, as images are comprised of discrete pixels. For Sobel method following two 3 x 3 kernels are used, one detects vertical edges and other horizontal edges.
\begin{equation}
 G_x = \begin{bmatrix}
       -1 & -2 & -1         \\[0.3em]
       0 & 0 & 0 \\[0.3em]
       1 & 2 & 1
     \end{bmatrix}*\Gamma \hspace{18pt}
 G_y = \begin{bmatrix}
       -1 & 0 & -1         \\[0.3em]
       -2 & 0 & 2 \\[0.3em]
       -1 & 0 & 1
     \end{bmatrix}*\Gamma
\end{equation}

where, $\Gamma$ is the image in which the edge has to detected. Prewitt edge detection uses a similar gradient method but with slightly different kernels. The magnitude and direction of the edge from Sobel anf Prewitt methods, can be obtained by,
\begin{equation}
G = \sqrt{G_x^2 + G_y^2} \hspace{20pt}
\Theta = arctan(G_y, G_x)
\label{eq:edge}
\end{equation}

An advanced technique used method which accounts for noise content and edge characteristics is Laplacian of Gaussian (LoG), in which regions of rapid intensity changes are highlighted by the Laplacian of the image. In this method, the image is first smoothed, generally by using a Gaussian kernel to reduce the sensitivity to noise. The Laplacian of the image with pixel intensity $I(x,y)$ is given by,

\begin{equation}
L(x,y) = \frac{\partial^2I}{\partial x^2} + \frac{\partial^2I}{\partial y^2}
\end{equation}

A smoothing filter has to be applied before the second derivative to reduce the effect of noise, LoG can be represented by a hybrid filter which incorporates both low pass filter and second derivative, Gaussian centered at 0 and standard deviation of $\sigma$ \cite{gon},

\begin{equation}
LoG(x,y) = - \frac{1}{\pi \sigma^4} \big [1 - \frac{x^2 + y^2}{2 \sigma^2} \big ]e^{- \frac{x^2 + y^2}{2 \sigma^2}}
\end{equation}

Canny edge detector is more complex algorithm which is superior to other edge detectors. The key objectives that underpins Canny method are, low, error rate, localized edge points, and single edge point response \cite{john}. In essence, Canny edge detector finds the optimal solution of the formulations of the aforementioned objectives. A two dimensional Gaussian function is used to reduce the noise, followed by gradient operation to calculate the strength and direction of edge at every point. Let, $f(x,y)$ be the input image and $G(x,y) = exp^{- \frac{x^2 + y^2}{2 \sigma^2}}$ denote the Gaussian function, the smoothed image is given by

\begin{equation}
f_s(x,y) = f(x,y)\star G(x,y)
\end{equation}

where, $\star$ denotes convolution. Further using equations $\ref{eq:grad}$ and $\ref{eq:edge}$, strength and direction of edges are calculated. A gradient method generally produces wide edges, nonmaxima suppression is used to thin out the edges.

Over the years, these techniques have been improved upon but they work better only in specific cases and still are generic and do not use the characteristics of the image. An improved Sobel method which uses soft-threshold wavelet de-noising combined with traditional Sobel technique but it works most efficiently on images with white Gaussian noise \cite{wen}. Another improved edge detection method uses a two level edge fusion model. On first level, typical edge detectors are used, where every pixel an edge score is calculated for every pixel \cite{xib}. Then results are combined at the second level using the Hadamard product with two additional edge estimations, based on edge spatial characteristics.

All the above techniques use gradient operator along with some smoothing filters and combination algorithms for edge detection. The major problem of these methods is that they are generic, so edge detected depends on noise, blurriness and illumination of the image. However, a method that and eliminates the process of pre-processing the image and depends on the features of the image being analysed is more likely to capture details in the image accurately. A self-learning algorithm which to adapts the filters and operators to the input image, irrespective of the class of image can provide the solution. In this paper, we have demonstrated an edge detection method based on image characteristics, using a dictionary constructed from the image itself. This method is different from traditional methods as it incorporates features of a input image, rather than on a pre-defined operators.


\section{The Dictionary based Edge Detection}
In recent years there has been a growing interest to study sparse representation of images using an overcomplete dictionary $\boldsymbol {D \in \mathbb{R}^{n \times K}}$ that contains $\boldsymbol K$ sample columns of a basis, $\boldsymbol {\{ d_j \}^K _{j=1}}$. The linear combination of these bases can be used to decompose a signal $ y \in \mathbb{R}^n$, or an image. The representation can be exact $\boldsymbol {y = Dx}$ or approximate, $\boldsymbol {\parallel y -Dx \parallel_p \leq \epsilon}$ depending of amount of noise, where 'p' can be 1,2 or $\boldsymbol \infty$, but generally $\boldsymbol p = 2$ is used. The accuracy of approximation depends upon the quality of dictionary. Methods such as basis pursuit and matching pursuit are used to construct dictionaries with reasonable approximation. Since, we are interested in the features of the input image and one of the best way to extract key features is principal component analysis.

\begin{figure}[h!]\centering
\begin{subfigure}[b]{.25\textwidth}
\centering
\includegraphics[keepaspectratio, width=\textwidth]{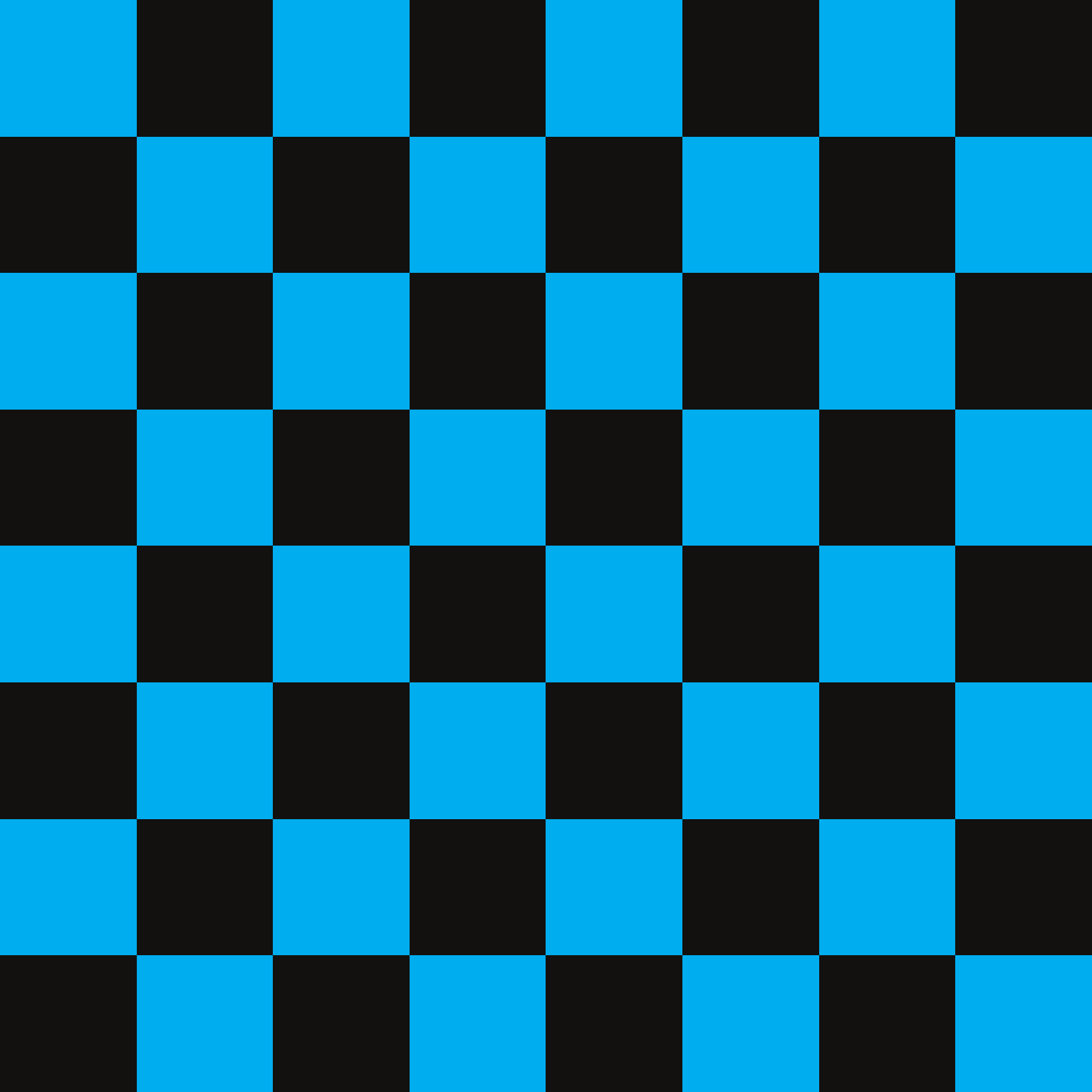}
\caption{}
\label{seg}
\end{subfigure}
\quad
\begin{subfigure}[p]{.3\textwidth}
\centering
\includegraphics[keepaspectratio, width=\textwidth]{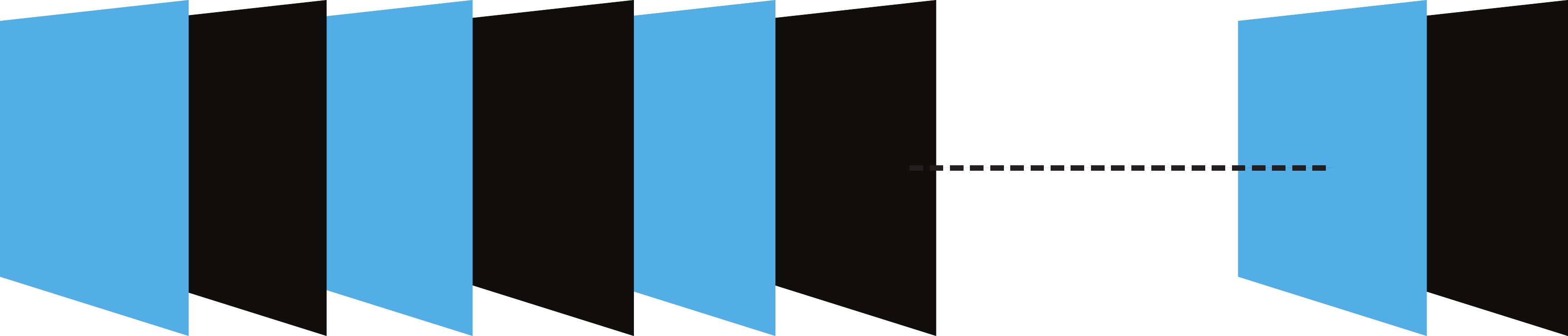}
\caption{}
\label{dict}
\end{subfigure}
\caption{Input Image: (a) Illustration of image $\bf \Gamma$ indicating small segments. (not to scale). (b) Illustration of 3D dictionary formed using small segments $(\bf s_i)$ of the image.}
\label{}
\end{figure}

Consider a test image, $\bf \Gamma$ of size NxN pixels (fig.\ref{seg}, test image has N = 512) in which color discontinuities, intensity variation are the key features for edge detection. 
In order to make the process more specific to an image, the filters have to be formed by analyzing the features of the image itself. The entire image is divided into small sections or sub-images ($\bf s_1, s_2, \cdots s_n$) of size n x n, (in this case n is 4, fig.\ref{dict}) to form the dictionary.

\begin{figure}[h!]\centering
\includegraphics[keepaspectratio, width=0.5\textwidth]{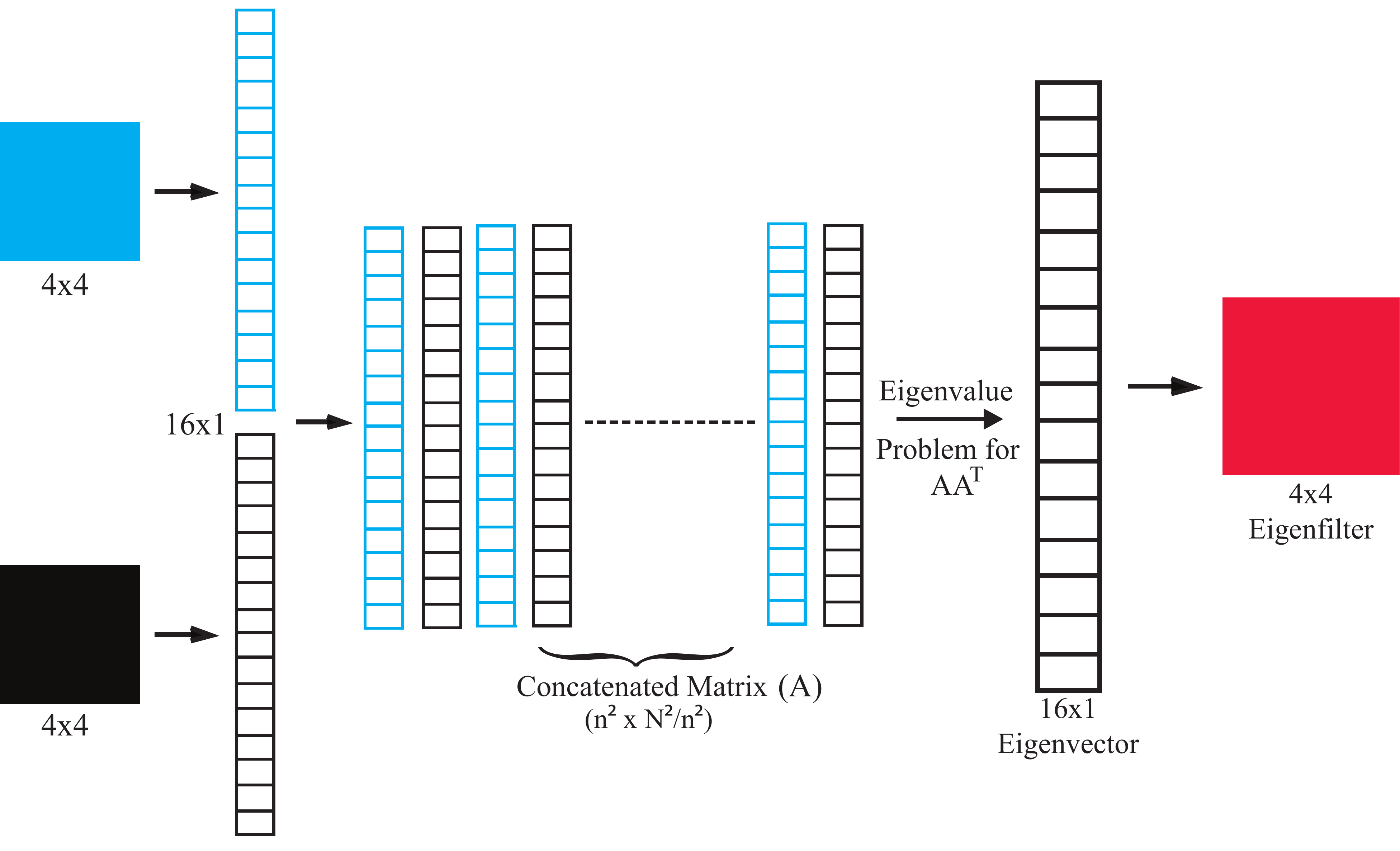}
\caption{Matrix formed by elements of dictionary.}
\label{mat}
\end{figure}

The choice of size of the segment cannot be arbitrary. The size of the segments is related to the kernels derived, which are used for detecting the edges. Optimal size of the segments is 3 x 3 to 5 x 5, if the size of segments is too small, kernels formed will not  be sufficient to capture all the important image characteristics. If the size is too large like 8 x 8, we observe that the image extracted  captures very less detail because key features of the image, is also interpreted as noise and hence is eliminated.

The dictionary is composed of $\bf s_1, s_2, \cdots s_n$, which are evidently images. All the elements of the dictionary are averaged, given by $\bf \Psi_s$, and subtracted from each element, $\bf \Phi_i$ to reduce the noise and increase the sharpness, due to low pass and high pass filtering nature of averaging and subtraction operations, respectively.

\begin{equation}
\boldsymbol \Psi_s = \frac{1}{n}\sum_{i=1}^n \boldsymbol s_i \hspace{45pt}
\bf \Phi_i = s_i - \Psi_s
\end{equation}

To find the key variations in each sub-image, principal components have to analyzed. But, the calculation of principal components for each sub-image involves solving of a large eigenvalue problem which may not be efficient to calculate. To get round this problem, we devised a method in which the n x n elements in the dictionary are transformed into column vectors of size $n^2$ x 1 and concatenated in matrix $\bf A$ (fig. \ref{mat}) of size $n^2$ x $N^2/n^2$. For the calculation in this paper, N = 512 and n = 4, hence the size of \textbf{A} is 16 x 16,384. The technique to construct the dictionary based on PCA, which is similar to the calculation of eigenfaces for recognition\cite{mat}. The eigenvectors of covariance matrix $\bf AA^T$ of size $n^2$ x $n^2$ (16 x 16) are calculated. This reduces the number of eigenvalue problems for each sub-image to just eigenvalue problem for one matrix. 

\begin{equation}
AA^T\overrightarrow{v_i} = \lambda_i \overrightarrow{v_i}
\end{equation}

\begin{equation}
\boldsymbol e_i = \boldsymbol \Gamma \ast \bf{u_i}
\end{equation}

\begin{figure}[h!]\centering
\begin{subfigure}[b]{.14\textwidth}
\centering
\includegraphics[keepaspectratio, width=\textwidth]{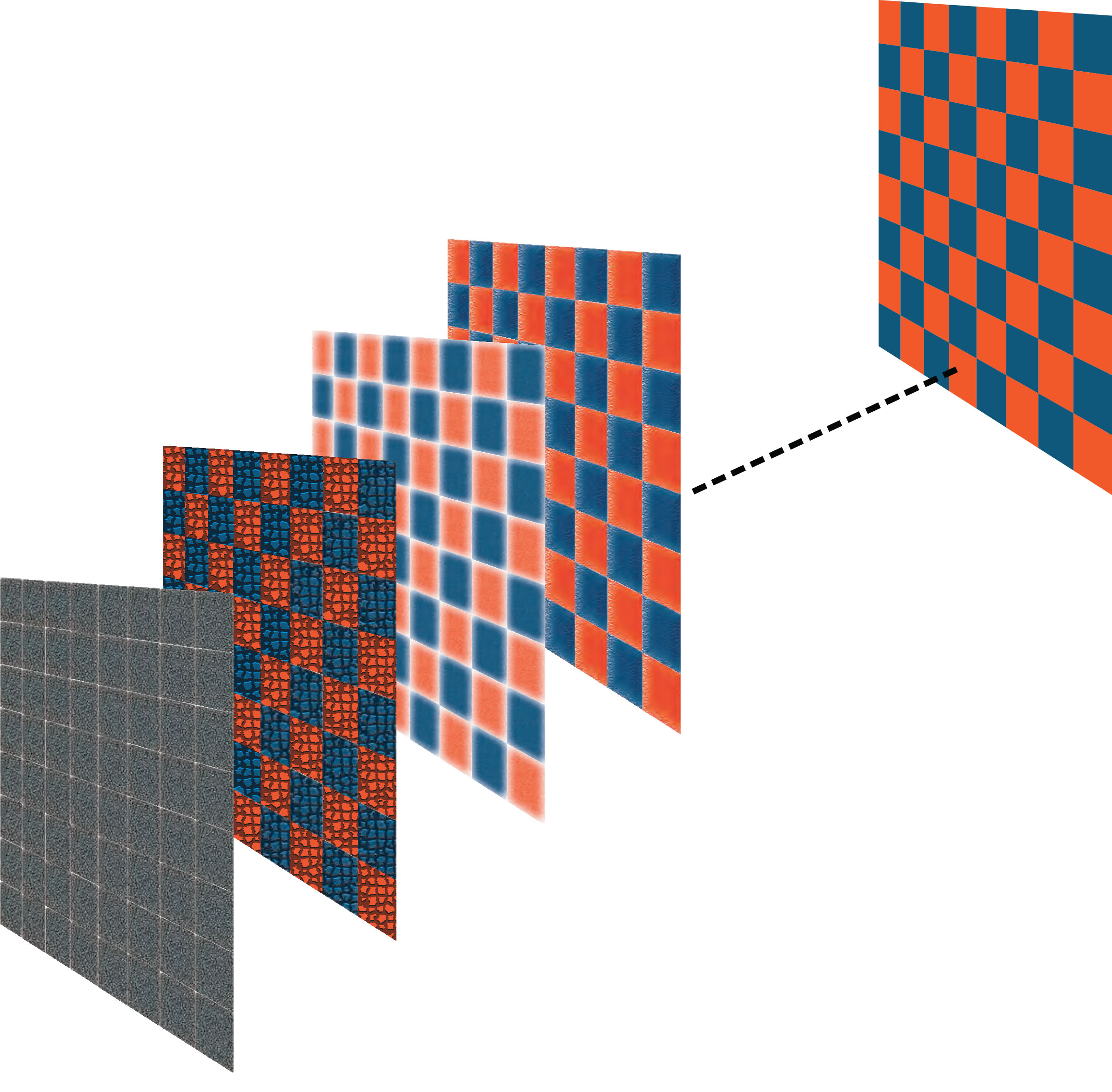}
\caption{}
\label{enh}
\end{subfigure}
\quad
\begin{subfigure}[b]{.14\textwidth}
\centering
\includegraphics[keepaspectratio, width=\textwidth]{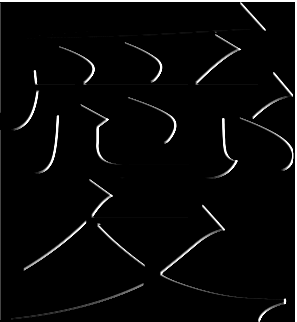}
\caption{}
\label{ef1}
\end{subfigure}
\quad
\begin{subfigure}[b]{.14\textwidth}
\centering
\includegraphics[keepaspectratio, width=\textwidth]{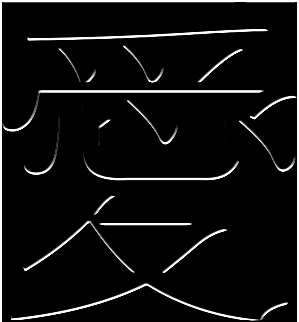}
\caption{}
\label{ef2}
\end{subfigure}
\caption{(a) 3d array of images $(\bf e_i)$ after convolution with $\bf \Gamma$ and thresholding. Intermediate edge images obtained after convolution with eigenfilter. (a) Edge image obtained by eigenfilter 14. (b) Edge image obtained by eigenfilter 15.}
\label{edge}
\end{figure}

Here, $\lambda_i$ are the eigenvalues and $\overrightarrow{v_i}$ are the eigenvectors, $\ast$ represents convolution of original image ($\boldsymbol \Gamma $) with $\bf u_i$, which are the rearranged eigenvectors to give  $\bf e_i$, intermediate edge image stored in a 3D array (fig. \ref{mat}). These $n^2$ eigenvectors of size $n^2$ x 1 (16 x 1) which are rearranged to form a dictionary of 16 kernels ($\bf u_i$) of size n x n, i.e 4 x 4 (fig.\ref{efil}). We have termed the derived kernels as eigenfilters. They have to be convoluted with image to detect edges along one direction, like a filter. The name eigenfilter originates from the work of Vaidyanathan and Nguyen \cite{vai}. After convolution of each eigenfilters with  $\bf \Gamma$, filtered images are stored stored in a 3D array (fig.\ref{enh}). The average $\bf \Psi_e$ of this 3D array is subtracted from individual elements of the array to give another 3D array $\boldsymbol \beta$.

\begin{figure}[h!]\centering

\end{figure}

\begin{equation}
\boldsymbol \Psi_e = \frac{1}{n}\sum_{i=1}^{16} \boldsymbol e_i \hspace{15pt}
\bf \Delta_i = e_i - \Psi_e
\end{equation}

The edge detection by individual eigenfilters is along one direction, as shown in fig. \ref{edge}. Edges detected by each of the eigenfilters for a binary object is shown in the appendix. To improve the quality of the edges, the edge images have to be combined. From edge image obtained, it is apparent that neighboring filters have similar characteristics. Thus, to enhance the edges, each pair of consecutive elements in the $\boldsymbol \beta$, maximum for each pixel in consecutive images is calculated, maximum of $\bf e_i$ and $\bf e_{i-1}$, as each consecutive eigenfilter is of similar characteristics (such as derivative filters) but in almost perpendicular direction (see fig. \ref{e1} and \ref{e2}, in appendix).

\begin{equation}
\boldsymbol \alpha_i = max(\bf e_i, e_{i-1})
\end{equation}

where, $\boldsymbol \alpha_i$ are stored in a 3D array $\boldsymbol \beta'$. The second last element of the new 3D array of 15 elements $\boldsymbol \beta'$ gives the desired edge because the last eigenfilter just reduces the intensity of the image (as shown in appendix) and is not useful for edge detection. The number of eigenfilter can vary depending on the size of sub-images taken and eigenvectors considered for convolution. 


\section{ Results and Analysis}
In this section, we have shown the application of the dictionary method in edge detection on various classes of image and provide solution to the problems with traditional methods. First, we have applied standard Sobel and Laplacian of Gaussian edge detection techniques on a Chinese character, which comprises of straight lines and curves and compared the results with dictionary based method. 

\begin{figure}[h!]\centering
\begin{subfigure}[b]{.10\textwidth}
\centering
\includegraphics[keepaspectratio, width=\textwidth]{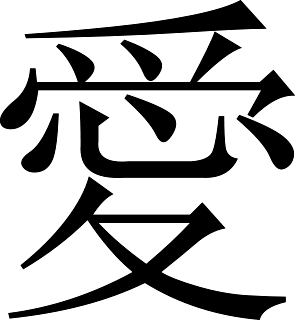}
\caption{}
\label{cdb}
\end{subfigure}
\quad
\begin{subfigure}[b]{.10\textwidth}
\centering
\includegraphics[keepaspectratio, width=\textwidth]{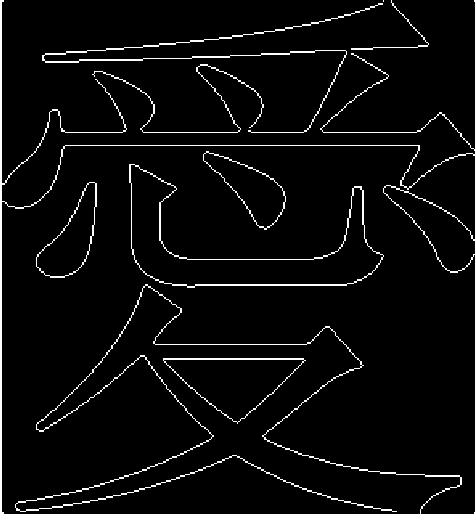}
\caption{}
\label{sob}
\end{subfigure}
\quad
\begin{subfigure}[b]{.10\textwidth}
\centering
\includegraphics[keepaspectratio, width=\textwidth]{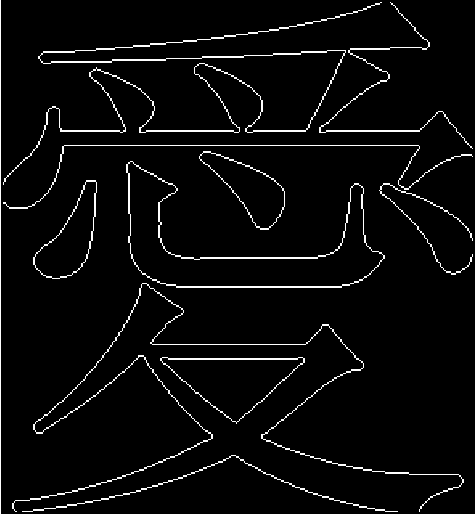}
\caption{}
\label{log}
\end{subfigure}
\quad
\begin{subfigure}[b]{.10\textwidth}
\centering
\includegraphics[keepaspectratio, width=\textwidth]{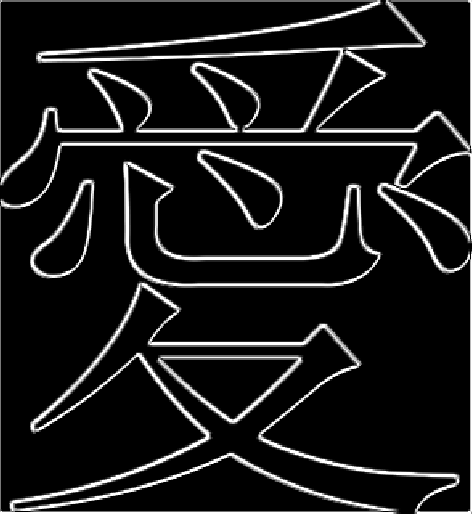}
\caption{}
\label{dict}
\end{subfigure}
\caption{Comparision with different methods of edge detection. The edges detected by dictionary based method are more detailed. (a) Original image of a Chinese character. (b) Edge detected by Sobel method. (c) Edge detected by LoG method. (d) Edge detected by dictionary based method.}
\label{}
\end{figure}

\begin{figure*}[t]\centering
\begin{subfigure}[b]{.31\textwidth}
\centering
\includegraphics[keepaspectratio, width=\textwidth]{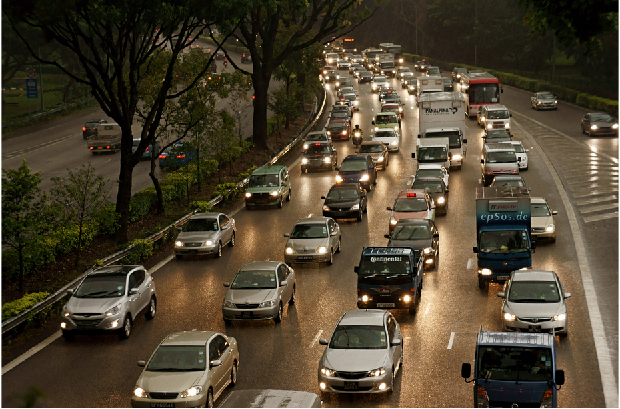}
\caption{}
\label{car}
\end{subfigure}
\quad
\begin{subfigure}[b]{.31\textwidth}
\centering
\includegraphics[keepaspectratio, width=\textwidth]{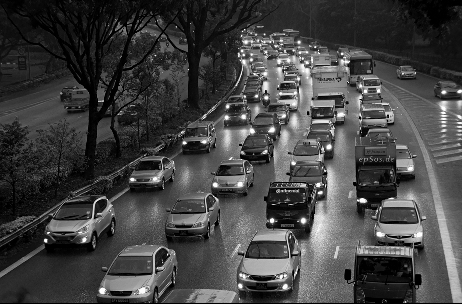}
\caption{}
\label{bcar}
\end{subfigure}
\quad
\begin{subfigure}[b]{.31\textwidth}
\centering
\includegraphics[keepaspectratio, width=\textwidth]{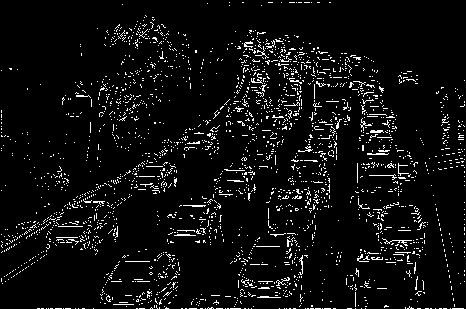}
\caption{}
\label{scar}
\end{subfigure}
\quad
\begin{subfigure}[b]{.31\textwidth}
\centering
\includegraphics[keepaspectratio, width=\textwidth]{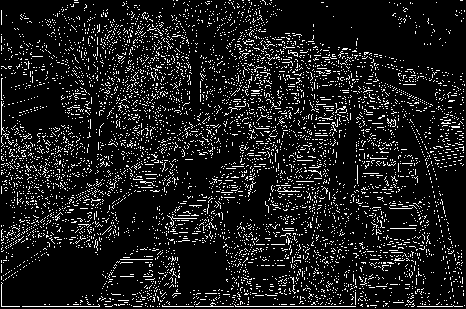}
\caption{}
\label{lcar}
\end{subfigure}
\quad
\begin{subfigure}[b]{.31\textwidth}
\centering
\includegraphics[keepaspectratio, width=\textwidth]{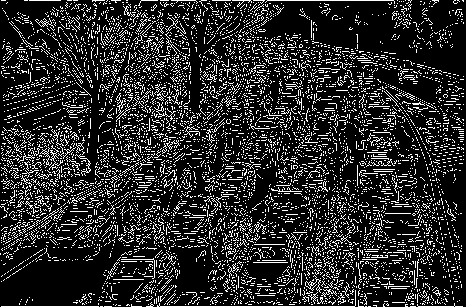}
\caption{}
\label{ccar}
\end{subfigure}
\quad
\begin{subfigure}[b]{.31\textwidth}
\centering
\includegraphics[keepaspectratio, width=\textwidth]{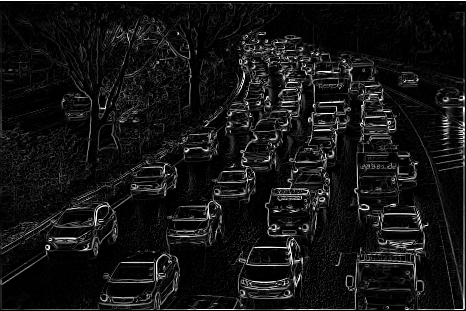}
\caption{}
\label{dcar}
\end{subfigure}
\caption{Edge detection in traffic image. (a) Original Image. (b) Grayscale image. (c) Edge detected by Sobel method. (d) Laplacian of Gaussian (e) Canny method (f) Dictionary based method.}
\label{}
\end{figure*}

The edge found by the dictionary based method is much more pronounced than the standard methods for simple images. To get the same performance in terms of edge continuity from the traditional techniques, the edge image has to be further subjected to morphological image processing techniques, such as dilation. The difference in the edges detected can be observed from the following cases.

\begin{figure}[h!]\centering
\begin{subfigure}[b]{.14\textwidth}
\centering
\includegraphics[keepaspectratio, width=\textwidth]{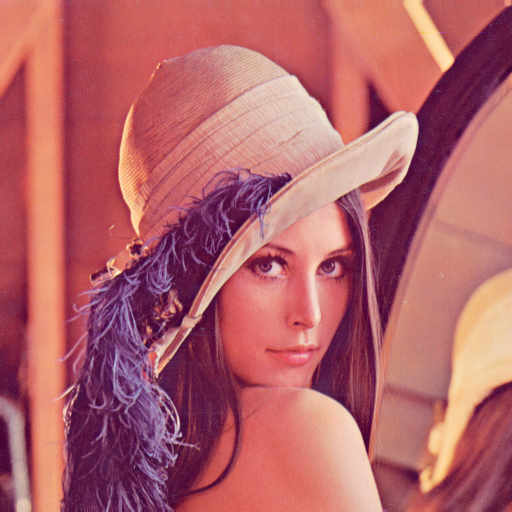}
\caption{}
\label{len}
\end{subfigure}
\quad
\begin{subfigure}[b]{.14\textwidth}
\centering
\includegraphics[keepaspectratio, width=\textwidth]{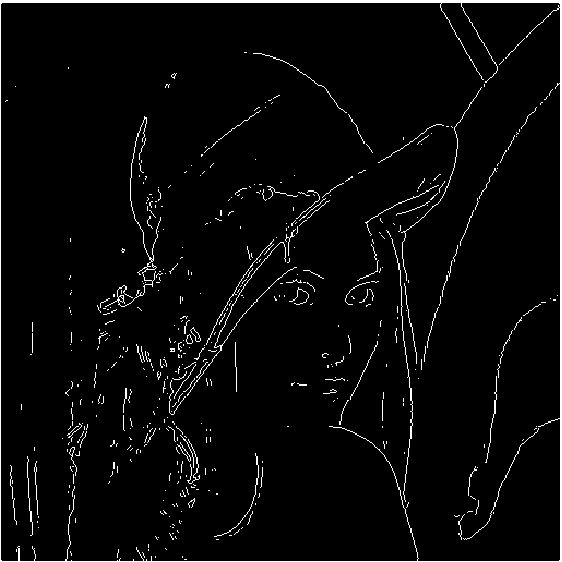}
\caption{}
\label{prew}
\end{subfigure}
\quad
\begin{subfigure}[b]{.14\textwidth}
\centering
\includegraphics[keepaspectratio, width=\textwidth]{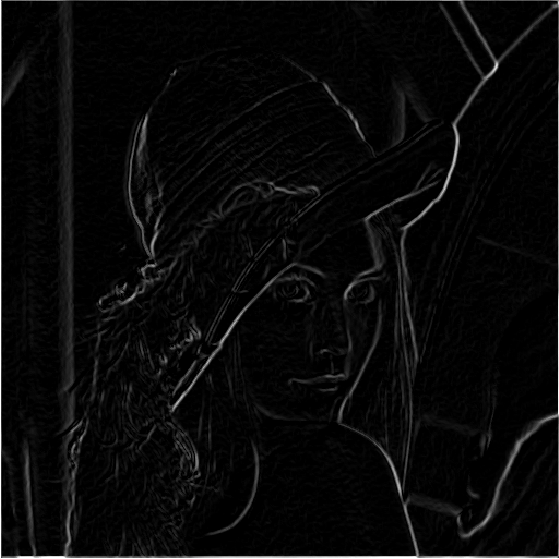}
\caption{}
\label{dict}
\end{subfigure}
\caption{Comparision between Sobel and Dictionary method of edge detection. (a) Original image of Lena. (b) Edge by Prewitt method. (c) Dictionary based method.}
\label{}
\end{figure}

For complex images such as the Lena image shown in fig. \ref{len} the output from dictionary based method is well defined, capture more features of the image and produces a continuous edge image as compared to Prewitt method. Also, for images which are not completely focused, the new algorithm performs much better. The calculation of eigenfilter by dividing into small segments increases the probability of capturing small peaks. The subtraction of $\bf \Psi_e$, reduces the remaining noise, improves edges as well as suppresses the background. The new dictionary based gives better results for every class of images such as text, face or cell images than the existing techniques, specially when the image is not captured at the highest resolution.

In practical situations, image is contaminated with noise, varying illumination shadows, glare etc. and these affect the quality of edges detected. Here, we have demonstrated application of various edge detection techniques on an images of roads full of cars (see fig. \ref{car}). The image is first converted to grayscale image (fig. \ref{bcar}) and the same image is used for all the methods without any pre-processing. Edges images from Sobel. Laplacian of Gaussian, Canny and Dictionary based method has been shown in fig. \ref{scar}, \ref{lcar}, \ref{ccar} and \ref{dcar}, respectively. It is quite evident from the results that dictionary based method procudes the best results with pre-processing. It also takes care of the varying illumination, noise, glare from headlights of the car and shadow which are shown as edges in output from LoG and Canny as they are more sensitive than Sobel method. This shows the advantage of having an algorithm that adapts to the input image and uses its feature for edge detection rather than a generic operator.


Accurate and fast edge detection techniques have many applications in image processing. In the following sub-sections, we will demonstrate the application and advantages of dictionary based method in barcode reading and cell counting.

\subsection{Barcode Reading} 
Barcodes have been used to store information for a long time. Primitive 1D barcodes are now being replaced by 2D barcodes which can contain much more data. But for reading these codes special hardware is required. Also, techniques to read 1D barcodes or stacked 2D barcodes such as, second derivative, peak location and EM algorithm require high resolution of scanned image \cite{hao}. In recent years, new smart-phone based reading barcode is gaining popularity. But still, for camera based interpretation of barcode is still limited, as binarizing the image to compute the bit pattern is not accurate in cases where image is blurred or distorted. To read the a 2D barcode segmentation of image, size detection and coordinates of four corners is required \cite{hao}. With accurate edge detection, the process of segmentation, size detection and corner locations can be enhanced. The edge image can also be used to calculate the rotation or tilt in the image more precisely.

\begin{figure}[h!]\centering
\begin{subfigure}[b]{.14\textwidth}
\centering
\includegraphics[keepaspectratio, width=\textwidth]{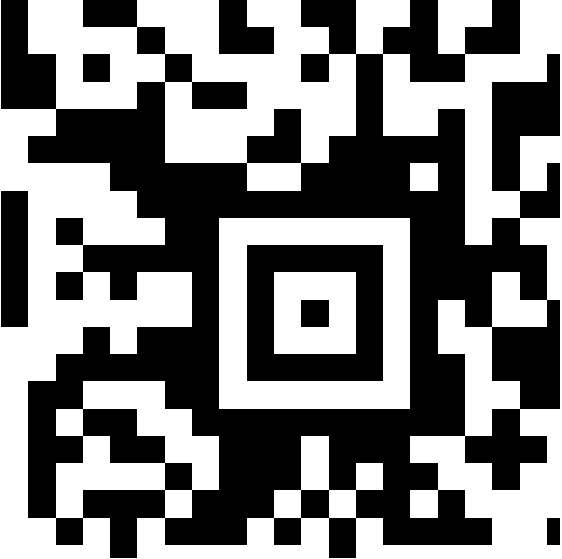}
\caption{}
\label{2d2}
\end{subfigure}
\quad
\begin{subfigure}[b]{.14\textwidth}
\centering
\includegraphics[keepaspectratio, width=\textwidth]{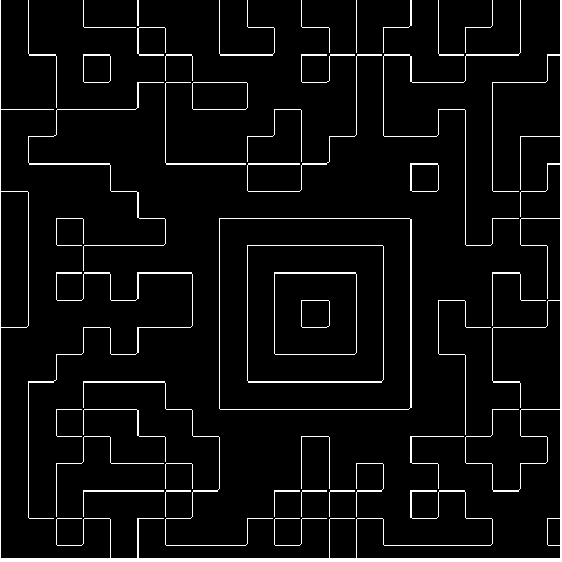}
\caption{}
\label{2dso}
\end{subfigure}
\quad
\begin{subfigure}[b]{.14\textwidth}
\centering
\includegraphics[keepaspectratio, width=\textwidth]{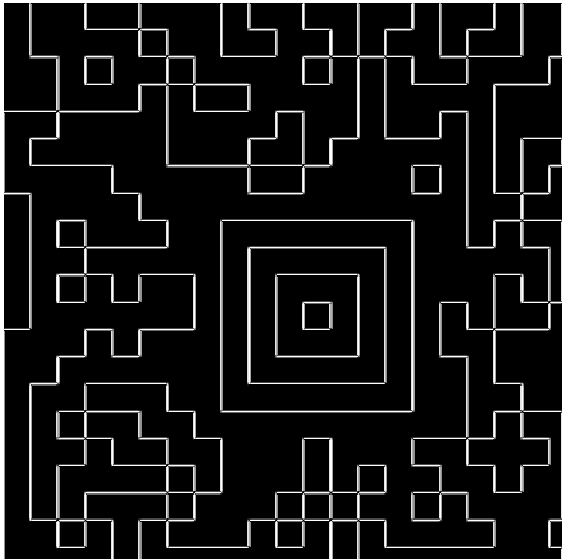}
\caption{}
\label{2db}
\end{subfigure}
\caption{Edge detection in highest resolution image of Aztec Code. (a) High quality Aztec Code image. (b) Edge detected by Sobel method. (c) Dictionary based method.}
\label{}
\end{figure}

As shown in fig. \ref{2d2}, we have a 2D barcode named Aztec Code. When edges are detected using Sobel and dictionary based methods the results are not very different for image with highest resolution. However, when the same image is blurred, as would happen in most practical cases, the advantage of self learning algorithm becomes evident. With the new algorithm which can detect edges from low resolution images, camera based interpretation of barcodes can be done more accurately using smart phones or web-cameras. 

\begin{figure}[h!]\centering
\begin{subfigure}[b]{.14\textwidth}
\centering
\includegraphics[keepaspectratio, width=\textwidth]{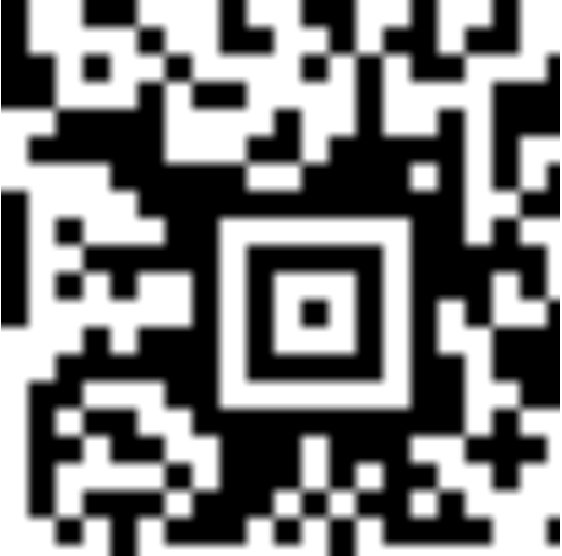}
\caption{}
\label{2db1}
\end{subfigure}
\quad
\begin{subfigure}[b]{.14\textwidth}
\centering
\includegraphics[keepaspectratio, width=\textwidth]{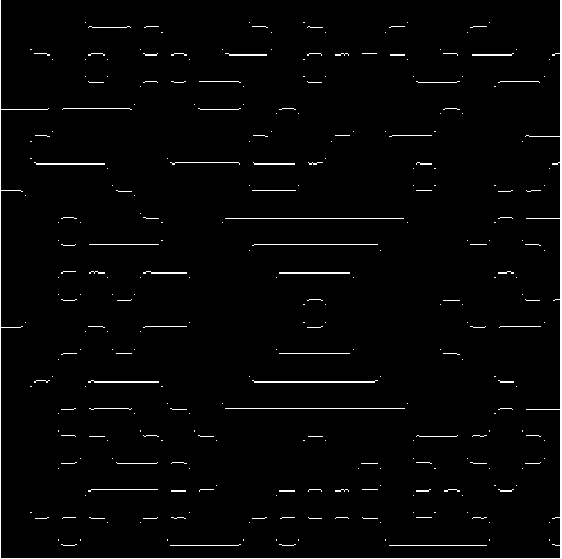}
\caption{}
\label{2dbso}
\end{subfigure}
\quad
\begin{subfigure}[b]{.14\textwidth}
\centering
\includegraphics[keepaspectratio, width=\textwidth]{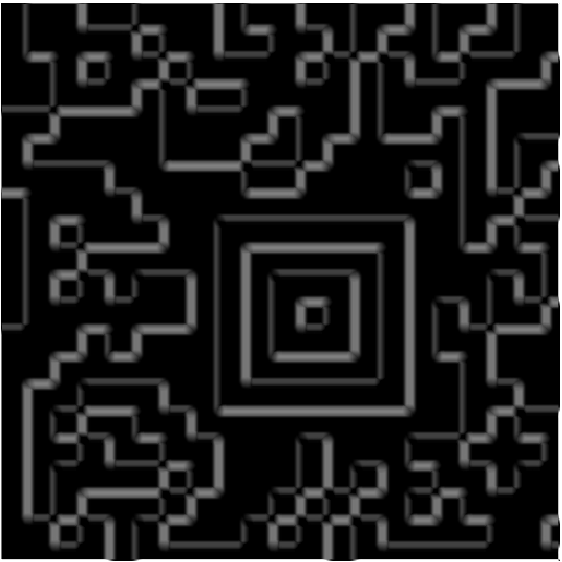}
\caption{}
\label{2dbarb}
\end{subfigure}
\caption{Edge detection in blurred image of Aztec Code,  blurred by averaging filter of size 12. (a) Blurred Aztec Code image. (b) Edge detected by Sobel method. (c) Dictionary based method.}
\label{}
\end{figure}

Another possible application can be in scanning of text images, where the edges are grainy or fonts are small to be captured clearly. Edge detection technique combined with average image capturing hardware can be used to produce high quality image. The output edge image from Prewitt method is not continuous as compared to output from dictionary method in the case where boundary of letters are not continuous and this could lead to errors in cases where small font size is used.

\subsection{Counting of Red Blood Cells}
In Biomedical imaging, edge detection algorithms can be used to count cells, detection of tumors etc. Counting of cells usually requires special apparatus which are expensive. For using techniques of snapshots of a sample to count and segment image, edge detection technique becomes extremely important.

\begin{figure}[h!]\centering
\begin{subfigure}[b]{.22\textwidth}
\centering
\includegraphics[keepaspectratio, width=\textwidth]{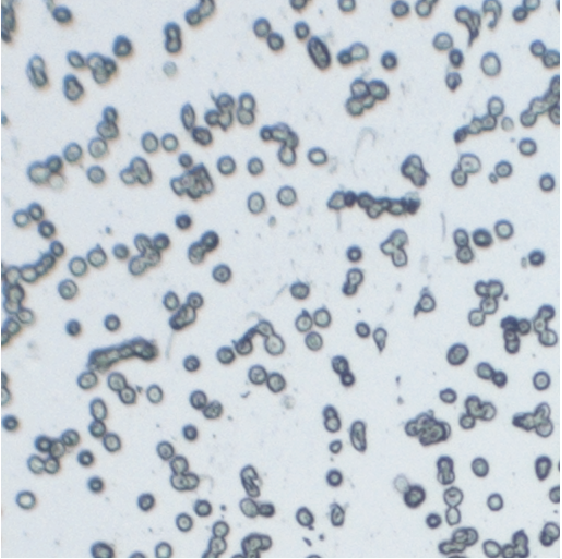}
\caption{}
\label{cel}
\end{subfigure}
\quad
\begin{subfigure}[b]{.22\textwidth}
\centering
\includegraphics[keepaspectratio, width=\textwidth]{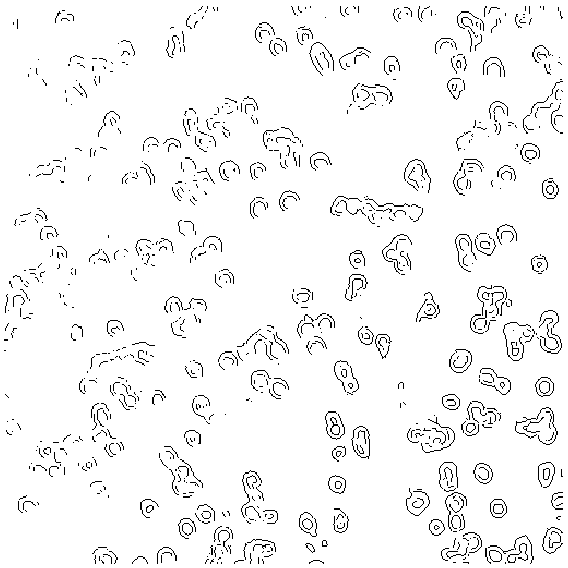}
\caption{}
\label{prew}
\end{subfigure}
\quad
\begin{subfigure}[b]{.22\textwidth}
\centering
\includegraphics[keepaspectratio, width=\textwidth]{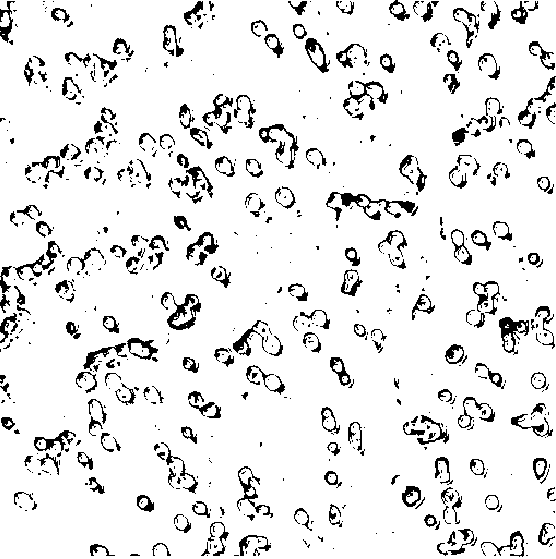}
\caption{}
\label{thr}
\end{subfigure}
\quad
\begin{subfigure}[b]{.22\textwidth}
\centering
\includegraphics[keepaspectratio, width=\textwidth]{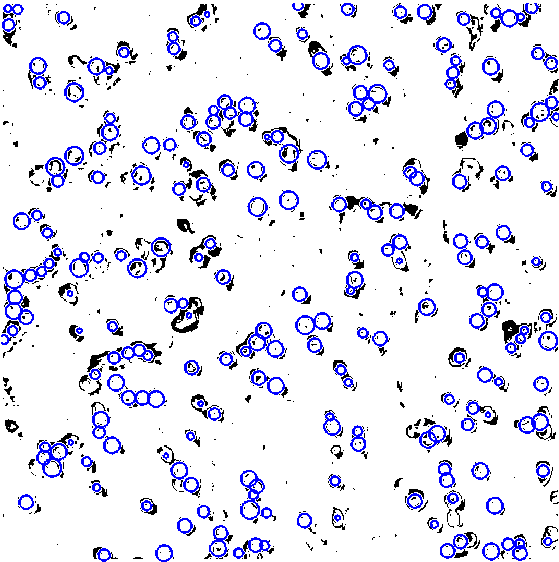}
\caption{}
\label{prew}
\end{subfigure}
\caption{Application on red blood cells. (a) Original RBC sample. (b) Edge detected by Sobel method. (c) Edge from Dictionary based method. (d) Estimation of number of RBC.}
\label{}
\end{figure}

With the use of the method described in this paper, it is possible to identify edges more accurately and in much more detail. In this paper, we have used an image of RBC cells taken from microscope with objective of 10x magnification(fig.\ref{cel}). 
%

The final image has dark background which has to be removed to make the edges more clearer and display in familiar form. For removing the background, a threshold on the intensity values of the pixels is applied to remove background. It is fairly simple, threshold varies slightly  for different images and can be estimated from the histogram of the image. After applying the threshold, advantage of proposed method over the existing techniques become evident.

While for Sobel method the edge detected are fainter and discontinuous. The dictionary based method is able to detect very faint boundaries and also of cells which are in the process of division. Sobel method is unable to detect cells that are joined partially or completely and in some cases it has not picked up the cells at all.

\begin{figure}[h!]\centering
\centering
\includegraphics[keepaspectratio, width=0.45\textwidth]{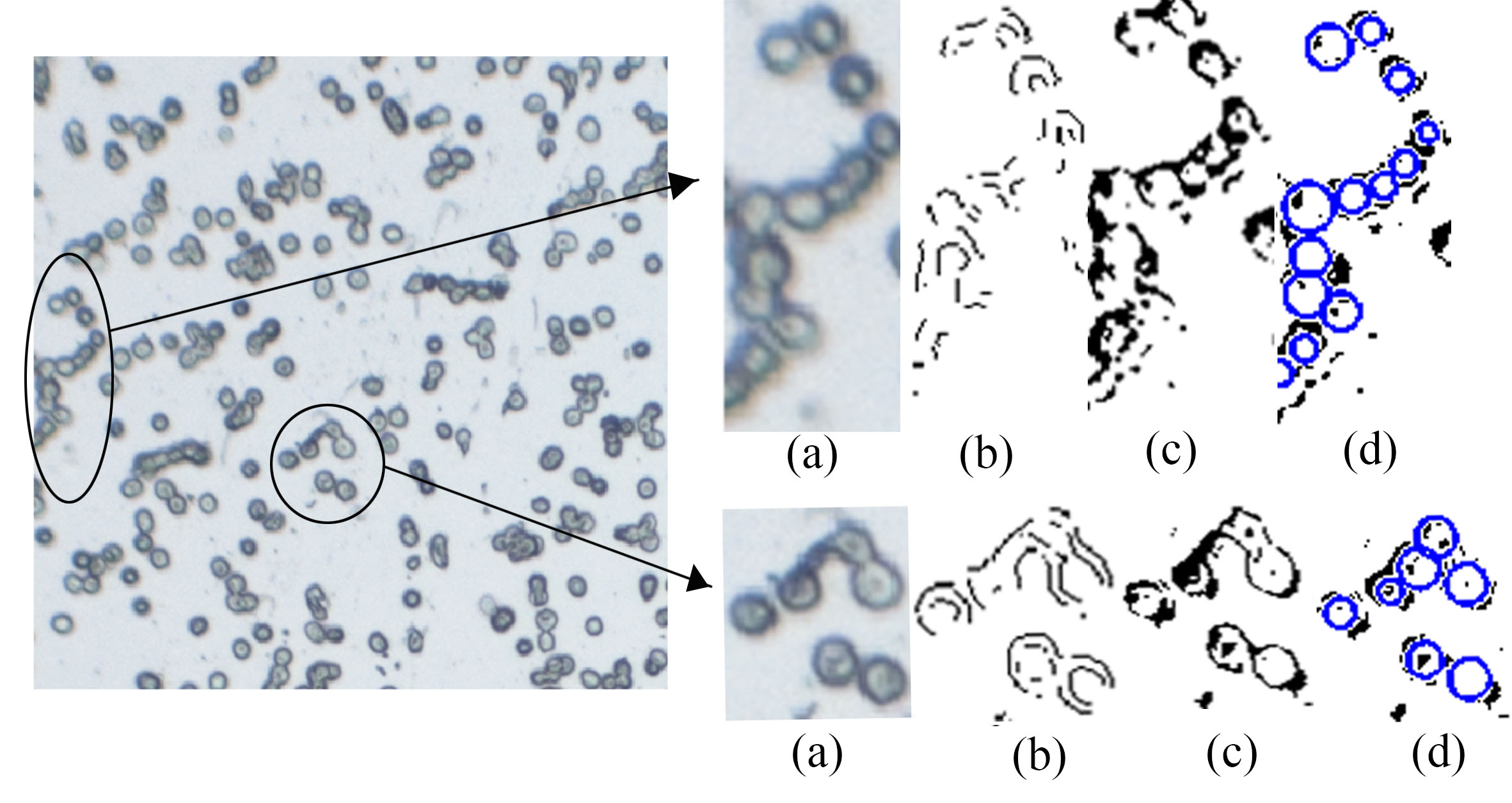}
\caption{(a) Magnified view of RBC sample; (b) Edge detected by Sobel method; (c) Edge detected by Dictionary based method; (d) Estimation of number of cell by approximating cells to a circle. Hence, better detection of edges can lead to better approximation and more accuracy of counting.}
\label{co}
\end{figure}

In fig. \ref{co}, a magnified view of two sections of image and corresponding edges detected by Sobel and dictionary methods has been presented. In first, the cells are joined together to form a long string and in second, there is a cluster of cells of different sizes and present at varying spacing. These are two types of sections in which estimating the boundary of cells and subsequently estimating the number of cells can be most difficult. As seen from the figure, dictionary based method produces a better edge image than Sobel method. To estimate the number of cells, we have taken the advantage of circular shape of the red blood cells and fitted the edge image with circles using circular Hough transform based algorithm which is also available in Matlab \cite{ath} \cite{yue}. This approach is used because of its performance in presence of noise and varying illumination. The radius of circles fitted has slight variation to account to cells in different stages. The range of radius of circle depends on the magnification of original image and typical cells size which is obtained from edge image. Another point to consider is that not all cells are circular and cells which are in the process of division look like two joined or half circles. Inter cellular space can also have similar circular features that can be picked in number estimation. From fig. \ref{co}d, it can be seen that fitting depends on the quality of edges and hence the accuracy of count.

Thus, a few precautions that should be taken in fitting circles to count number of cells. In this paper, we have used range of radius from 3 to 8 pixels for both methods. From fig. \ref{co}, estimation by Sobel method is not accurate as it misses many cells, counts inter cellular spaces and size of the estimated circle is not close to the cell size. Excluding the cells that are not completely contained within the image, total number of cells are 226, and the cells number of cells detected by dictionary based method is 225, with an error of $0.44 \%$.

For applications such as counting of cells, the dictionary based method has advantage over other methods as it detects more accurate and continuous edges and allows fitting circles which not only counts the RBCs, but also estimates the average and variation of radii of the cells.

\section{Conclusion}
In this paper, we have proposed a new edge detection algorithm using dictionary comprised of the key features of the image. Since the eigenfilters are derived from the input image, it becomes image specific and sensitive to its features. This enables the algorithm to adapt to different classes of images and produce better results. Its shown in the paper, results from the new method is much more accurate and detailed than existing algorithms. 

At the beginning of the paper, we have highlighted major issues that exists with the traditional methods. In general, they reqires pre-processing of the image and it depends on the type of noise, class of image etc. A method that takes the inputs image and used its features for calculation edges eliminates this step. Furthermore, dictionary based method demonstrated in this paper is not affected by noise, blurriness, class of image or varying illumination, shown through various examples and provides a smooth solution to the problem of edge detection.

We have demonstrated its application in variety of areas and compared the results to that of existing algorithms. It can detect accurate edges from blurry image, enabling accurate image segmentation thus making barcode reading using cameras more reliable. It also improves the quality of text in cases where letters are grainy with small fonts. Dictionary based method can also be used to count the number of red blood cells in a sample without the need for specialized equipments.


%

\newpage
\appendices
\section{Eigenfilters and Corresponding Edge Images}
\begin{figure}[h!]\centering
\begin{subfigure}[b]{0.10\textwidth}
\centering
\includegraphics[keepaspectratio, width=\textwidth]{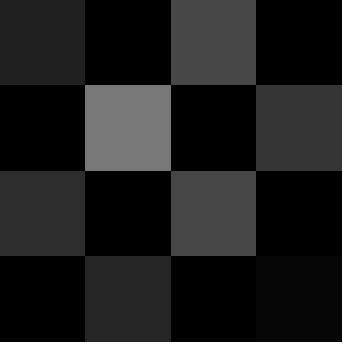}
\caption{Eigenfilter $~$ $\# 1$}
\end{subfigure}
\quad
\begin{subfigure}[b]{0.09\textwidth}
\centering
\includegraphics[keepaspectratio, width=\textwidth]{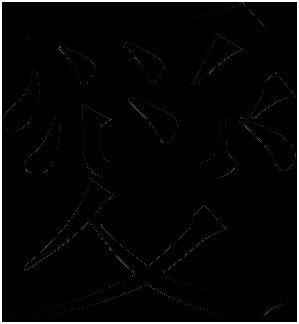}
\caption{Edge $~$ $\# 1$}
\end{subfigure}
\quad
\begin{subfigure}[b]{.10\textwidth}
\centering
\includegraphics[keepaspectratio, width=\textwidth]{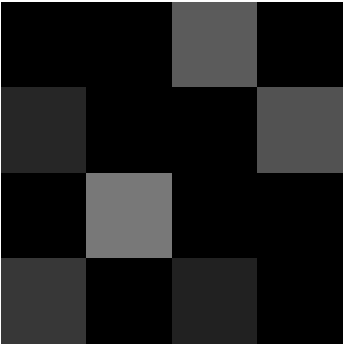}
\caption{Eigenfilter $~$ $\# 2$}
\end{subfigure}
\quad
\begin{subfigure}[b]{0.09\textwidth}
\centering
\includegraphics[keepaspectratio, width=\textwidth]{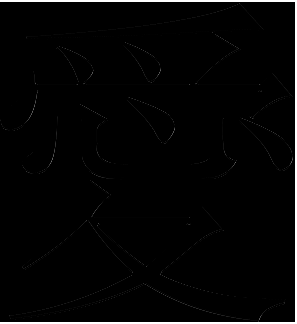}
\caption{Edge $~$ $\# 2$}
\end{subfigure}
\quad
\begin{subfigure}[b]{.10\textwidth}
\centering
\includegraphics[keepaspectratio, width=\textwidth]{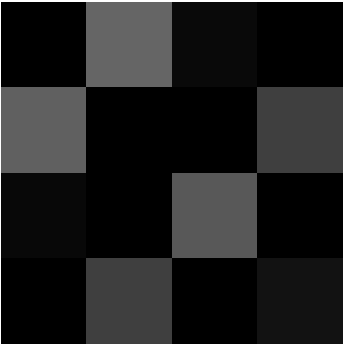}
\caption{Eigenfilter $~$ $\# 3$}
\end{subfigure}
\quad
\begin{subfigure}[b]{0.09\textwidth}
\centering
\includegraphics[keepaspectratio, width=\textwidth]{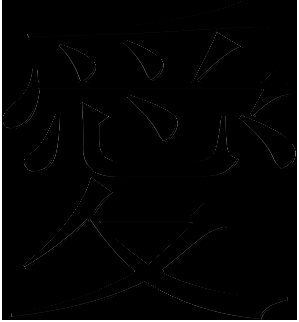}
\caption{Edge $~$ $\# 3$}
\end{subfigure}
\quad
\begin{subfigure}[b]{.10\textwidth}
\centering
\includegraphics[keepaspectratio, width=\textwidth]{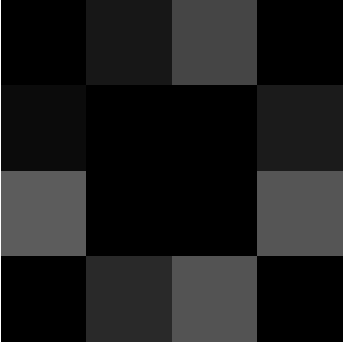}
\caption{Eigenfilter $\# 4$}
\end{subfigure}
\quad
\begin{subfigure}[b]{0.09\textwidth}
\centering
\includegraphics[keepaspectratio, width=\textwidth]{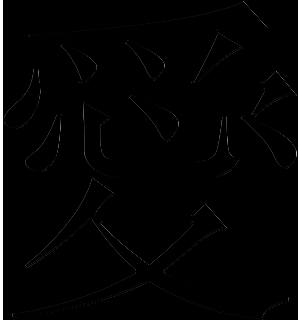}
\caption{Edge $~$ $\# 4$}
\end{subfigure}
\quad
\begin{subfigure}[b]{.10\textwidth}
\centering
\includegraphics[keepaspectratio, width=\textwidth]{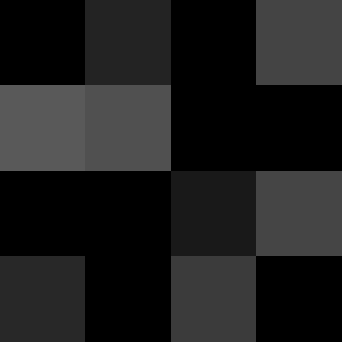}
\caption{Eigenfilter $~$ $\# 5$}
\end{subfigure}
\quad
\begin{subfigure}[b]{0.09\textwidth}
\centering
\includegraphics[keepaspectratio, width=\textwidth]{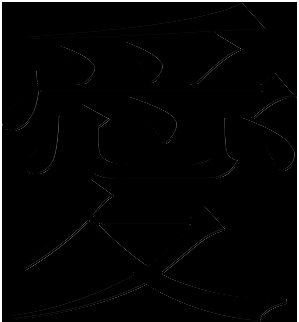}
\caption{Edge $~$ $\# 5$}
\end{subfigure}
\quad
\begin{subfigure}[b]{.10\textwidth}
\centering
\includegraphics[keepaspectratio, width=\textwidth]{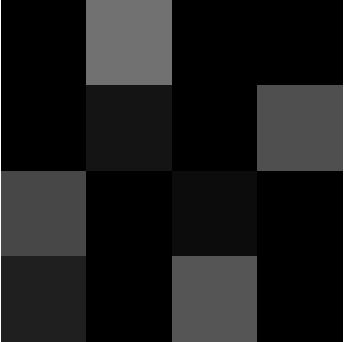}
\caption{Eigenfilter $~$ $\# 6$}
\end{subfigure}
\quad
\begin{subfigure}[b]{0.09\textwidth}
\centering
\includegraphics[keepaspectratio, width=\textwidth]{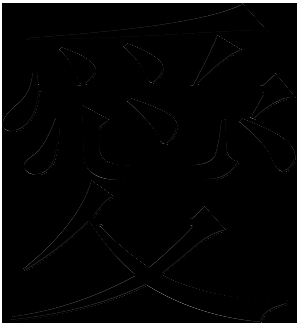}
\caption{Edge $~$ $\# 6$}
\end{subfigure}
\quad
\begin{subfigure}[b]{.10\textwidth}
\centering
\includegraphics[keepaspectratio, width=\textwidth]{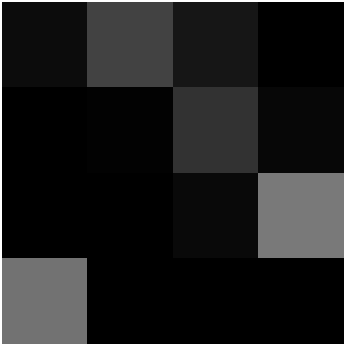}
\caption{Eigenfilter $~$ $\# 7$}
\end{subfigure}
\quad
\begin{subfigure}[b]{0.09\textwidth}
\centering
\includegraphics[keepaspectratio, width=\textwidth]{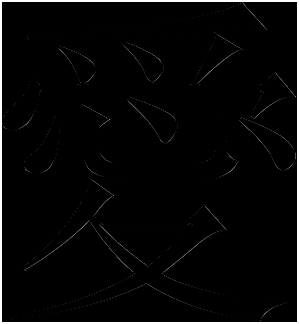}
\caption{Edge $~$ $\# 7$}
\end{subfigure}
\quad
\begin{subfigure}[b]{.10\textwidth}
\centering
\includegraphics[keepaspectratio, width=\textwidth]{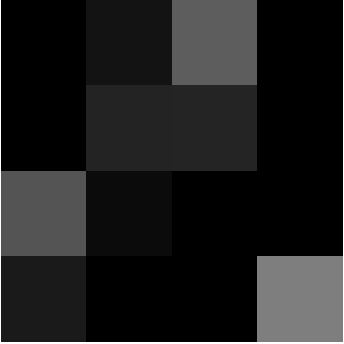}
\caption{Eigenfilter $~$ $\# 8$}
\end{subfigure}
\quad
\begin{subfigure}[b]{0.09\textwidth}
\centering
\includegraphics[keepaspectratio, width=\textwidth]{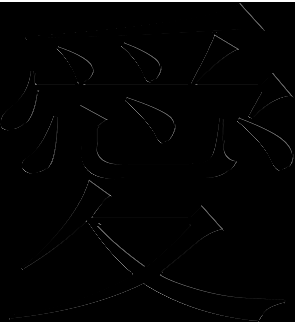}
\caption{Edge $~$ $\# 8$}
\end{subfigure}
\quad
\begin{subfigure}[b]{.10\textwidth}
\centering
\includegraphics[keepaspectratio, width=\textwidth]{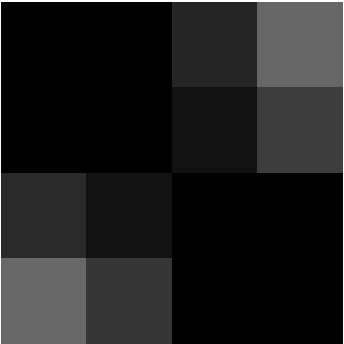}
\caption{Eigenfilter $~$ $\# 9$}
\end{subfigure}
\quad
\begin{subfigure}[b]{0.09\textwidth}
\centering
\includegraphics[keepaspectratio, width=\textwidth]{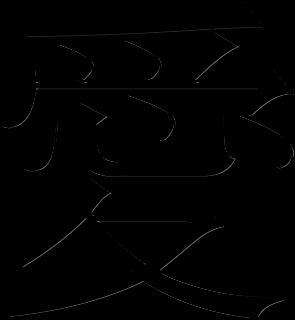}
\caption{Edge $~$ $\# 9$}
\end{subfigure}
\quad
\begin{subfigure}[b]{.10\textwidth}
\centering
\includegraphics[keepaspectratio, width=\textwidth]{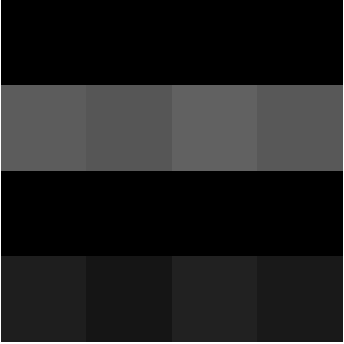}
\caption{Eigenfilter $~$ $\# 10$}
\end{subfigure}
\quad
\begin{subfigure}[b]{0.09\textwidth}
\centering
\includegraphics[keepaspectratio, width=\textwidth]{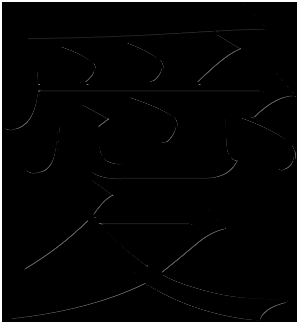}
\caption{Edge $~$ $\# 10$}
\end{subfigure}
\quad
\begin{subfigure}[b]{.10\textwidth}
\centering
\includegraphics[keepaspectratio, width=\textwidth]{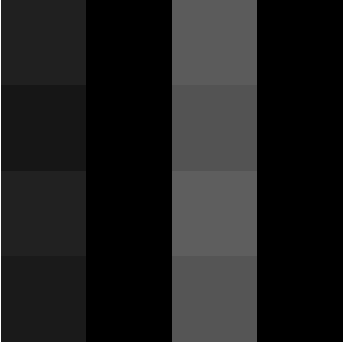}
\caption{Eigenfilter $~$ $\# 11$}
\end{subfigure}
\quad
\begin{subfigure}[b]{0.09\textwidth}
\centering
\includegraphics[keepaspectratio, width=\textwidth]{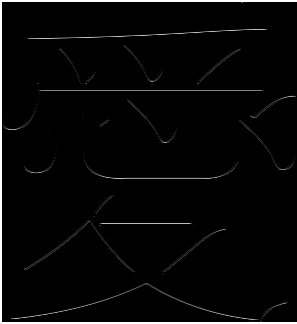}
\caption{Edge $~$ $\# 11$}
\end{subfigure}
\quad
\begin{subfigure}[b]{.10\textwidth}
\centering
\includegraphics[keepaspectratio, width=\textwidth]{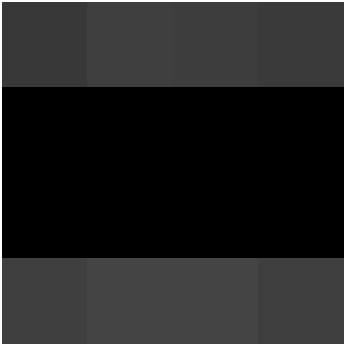}
\caption{Eigenfilter $~$ $\# 12$}
\end{subfigure}
\quad
\begin{subfigure}[b]{0.09\textwidth}
\centering
\includegraphics[keepaspectratio, width=\textwidth]{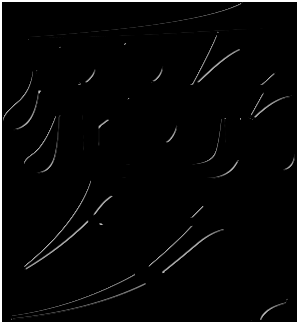}
\caption{Edge $~$ $\# 12$}
\end{subfigure}
\caption{Eigenfilter displayed as an image with respective edge images (before combining for final edge image).}
\end{figure}

\begin{figure}[t]
\begin{subfigure}[b]{.10\textwidth}
\centering
\includegraphics[keepaspectratio, width=\textwidth]{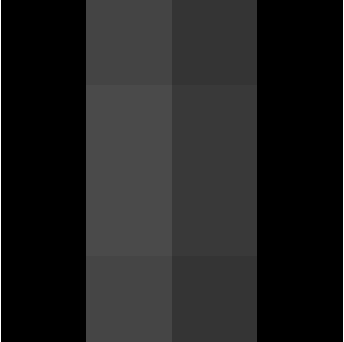}
\caption{Eigenfilter $~$ $\# 13$}
\end{subfigure}
\quad
\begin{subfigure}[b]{0.09\textwidth}
\centering
\includegraphics[keepaspectratio, width=\textwidth]{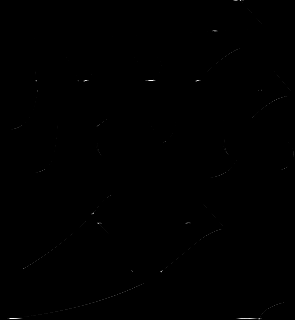}
\caption{Edge $~$ $\# 13$}
\end{subfigure}
\quad
\begin{subfigure}[b]{.10\textwidth}
\centering
\includegraphics[keepaspectratio, width=\textwidth]{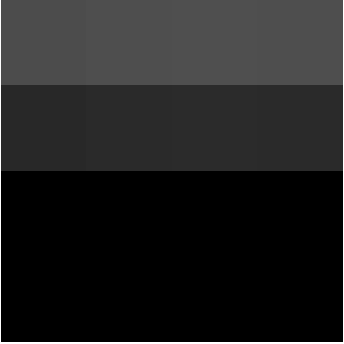}
\caption{Eigenfilter  $\# 14$}
\label{e1}
\end{subfigure}
\quad
\begin{subfigure}[b]{0.09\textwidth}
\centering
\includegraphics[keepaspectratio, width=\textwidth]{c14.png}
\caption{Edge $~$ $\# 14$}
\end{subfigure}
\quad
\begin{subfigure}[b]{.10\textwidth}
\centering
\includegraphics[keepaspectratio, width=\textwidth]{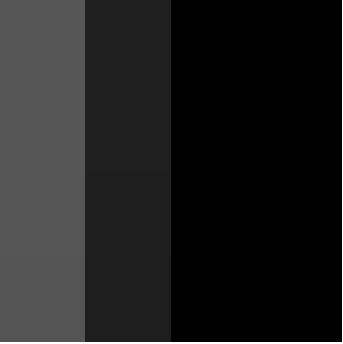}
\caption{Eigenfilter $~$ $\# 15$}
\label{e2}
\end{subfigure}
\quad
\begin{subfigure}[b]{0.09\textwidth}
\centering
\includegraphics[keepaspectratio, width=\textwidth]{c15.png}
\caption{Edge $~$ $\# 15$}
\end{subfigure}
\quad
\begin{subfigure}[b]{.10\textwidth}
\centering
\includegraphics[keepaspectratio, width=\textwidth]{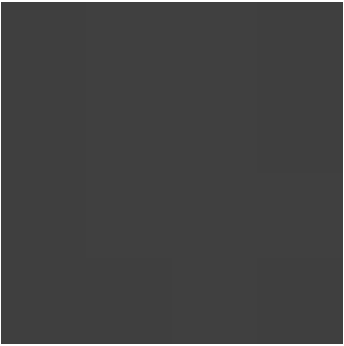}
\caption{ Eigenfilter $~$ $\# 16$}
\end{subfigure}
\quad
\begin{subfigure}[b]{0.09\textwidth}
\centering
\includegraphics[keepaspectratio, width=\textwidth]{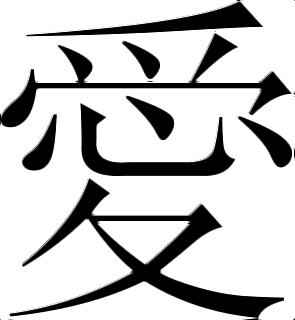}
\caption{Edge $~$ $\# 16$}
\end{subfigure}
\caption{Eigenfilter displayed as an image with respective edge images (before combining for final edge image).}
\label{}
\end{figure}
%
%


\ifCLASSOPTIONcaptionsoff
  \newpage
\fi

\end{document}